\documentclass{article}
\usepackage{ijcai19}

\usepackage{times}
\usepackage{soul}
\usepackage{url}
\usepackage[hidelinks]{hyperref}
\usepackage[utf8]{inputenc}
\usepackage[small]{caption}
\usepackage{graphicx}
\graphicspath{ {./images/} }
\usepackage{amsmath}
\usepackage{booktabs}
\usepackage{algorithm}
\usepackage{algorithmic}
\usepackage{enumitem}
\DeclareMathOperator*{\argmin}{argmin}
\usepackage{amssymb}
\usepackage{adjustbox}

\title{Autoencoder-Based Incremental Class Learning without Retraining on Old Data}

\author{
Euntae Choi$^1$
\and
Kyungmi Lee$^2$\and
Kiyoung Choi$^3$
\affiliations
$^{1,3}$Seoul National University\\
$^2$Massachusetts Institute of Technology\\
\emails
echoi@dal.snu.ac.kr,
kyungmi@mit.edu,
kchoi@snu.ac.kr
}

\begin{document}

\maketitle

\begin{abstract}
  Incremental class learning, a scenario in continual learning context where classes and their training data are sequentially and disjointedly observed, challenges a problem widely known as \textit{catastrophic forgetting}. In this work, we propose a novel incremental class learning method that can significantly reduce memory overhead compared to previous approaches. Apart from conventional classification scheme using softmax, our model bases on an autoencoder to extract prototypes for given inputs so that no change in its output unit is required. It stores only the mean of prototypes per class to perform metric-based classification, unlike rehearsal approaches which rely on large memory or generative model. To mitigate catastrophic forgetting, regularization methods are applied on our model when a new task is encountered. We evaluate our method by experimenting on CIFAR-100 and CUB-200-2011 and show that its performance is comparable to the state-of-the-art method with much lower additional memory cost.
\end{abstract}

\section{Introduction}

 Modern deep neural networks (DNNs) have made an unquestionable success in various fields, especially under offline setting where the task for a network is fixed and all training data are provided simultaneously. In many real-world applications, however, DNNs are also required to perform well under continual settings. For instance, there can be a large set of medical images collected over a long period, say, longer than 10 years and each image is deleted after some period due to privacy concerns. Even worse, some images may come from previously unseen classes. Unfortunately, if a DNN that has been trained offline is naively fine-tuned on new set of data, it loses most of the representations for old ones. This notorious phenomenon is widely known as \textit{catastrophic forgetting} (or \textit{interference}) \cite{mccloskey1989catastrophic,french1999catastrophic}. The most basic way to avoid forgetting is to mix both old and new data together to construct a new batch, and then re-train the network offline (\textit{joint training}). Considering large datasets and complex network architectures used nowadays, this solution requires too much computation and memory cost, making it unscalable.
 
 Researchers are paying great attention to the problem of catastrophic forgetting in Continual Learning(CL) and making progress with various approaches. They either extend loss function to prevent important parameters from drastic changes\cite{pmlr-v70-zenke17a}, use an external memory module to store a part exemplars of old tasks \cite{Rebuffi_2017_CVPR}, or employ a generative model to generate exemplars to remove additional memory cost \cite{kemker2018fearnet}. At the same time, there is also a controversy with the standard on how we should evaluate CL approaches and make fair comparisons \cite{maltoni2018continuous,hsu2018re}. This issue is raised because some methods like \cite{kirkpatrick2017overcoming,pmlr-v70-zenke17a} work quite well in a multi-task learning scenario, but show significant performance drop in incremental class learning (ICL) scenario \cite{hsu2018re,parisi2018continual}. 
 
We propose a novel ICL algorithm that stores only one code vector (mean) per class and does not retrain any other data from old tasks, thereby reducing memory and computation cost and alleviating the privacy issues simultaneously. The base of our model is an autoencoder which learns how to (1) find an important sub-manifold of data distribution and (2) make code vectors well separated in cosine metric space so that a metric-based classification method (the cosine-version of \textit{nearest-class-mean} (NCM) \cite{6517188}) works well. The autoencoder is first trained on a fixed number of classes and then Synaptic Intelligence (SI) \cite{pmlr-v70-zenke17a} or Memory Aware Synapses (MAS) \cite{10.1007/978-3-030-01219-9_9} loss is added during incremental training steps to resist forgetting. So basically, our model can be viewed as a combination of architectural and regularization methodology. Alternatively, this is an approach that makes previous regularization methodologies applicable to ICL context with minimal overhead.

In the rest of this paper, we first review numerous previous CL approaches in the \textit{Related Works}. We then explain the details of ours in the \textit{Methodology}. Experimental settings, results, and comparisons with previous works are presented in the \textit{Experiments}. Finally, we summarize and suggest possible future works in the \textit{Conclusion}.

\section{Related Works}

Recent studies on CL can be categorized into (1) \textit{regularization}, (2) \textit{rehearsal}, and (3) \textit{dual memory system} approaches.

\textit{Regularization Strategies} \cite{li2018learning,kirkpatrick2017overcoming,pmlr-v70-zenke17a,10.1007/978-3-030-01219-9_9,serra2018overcoming} focus on designing loss term to retain the representations for old tasks by e.g., prohibiting drastic update of important parameters. Learning without Forgetting (LwF) \cite{li2018learning} records the network's response to old tasks used in Knowledge Distillation \cite{44873} loss term and uses them to encourage the network to make similar prediction after being trained for new tasks. In Elastic Weight Consolidation (EWC) \cite{kirkpatrick2017overcoming}, it is assumed that keeping the value of loss function for old tasks low can lead to less forgetting. Therefore, each parameter's contribution (or importance) to the change in loss function is approximated as the diagonal of the Fisher information matrix. Then it is combined with quadratic penalty term on the parameter change to compose surrogate loss. Intuitively, EWC mitigates forgetting by forcing parameters that contributed a lot for old tasks to settle. SI \cite{pmlr-v70-zenke17a} is a variant of EWC which replaces the Fisher information matrix-based importance measure with the summation of the approximated change in loss function divided by the amount of update for each parameter, thus reducing computation cost. And an extra phase for computing the measure is spared because SI's measure can be computed online. Memory Aware Synapses(MAS) \cite{10.1007/978-3-030-01219-9_9} suggests using the averaged L2 norm of per-parameter gradient as the importance measure.

However, some papers \cite{hsu2018re,parisi2018continual} report that under ICL setting, recent regularization approaches have no clear advantage over simply fine-tuning the network. Compared to several rehearsal approaches \cite{Rebuffi_2017_CVPR,gepperth:hal-01418123}, the accuracy gap was significantly large (70\% for Split MNIST and 30\% for CUB-200). 


\textit{Rehearsal approaches} \cite{gepperth:hal-01418123,Rebuffi_2017_CVPR,wu2018incremental,NIPS2017_7225,kemker2018fearnet} store or generate old data to augment new training batches that helps the network retain its old representations. Notably, Incremental Classifier and Representation Learning (iCaRL) \cite{Rebuffi_2017_CVPR} is a mixture of \textit{rehearsal} and \textit{regularization} approaches. It has an external memory with fixed capacity to save previous exemplars that augment loss function with distillation term. Classification is done with \textit{nearest-mean-of-exemplars} rule, which uses the average of extracted feature vectors as the class mean instead of true class mean used in NCM \cite{6517188} classifier. To maintain memory capacity, the number of exemplars per class decreases by selecting those closest to their class mean after learning each task. While the memory helps iCaRL's strong performance, it still poses scalability problem as the number of classes grows. Incremental Classifier Learning with GAN (ICwGAN) \cite{wu2018incremental} does not use real exemplars as in iCaRL, but trains a GAN to get generated data (\textit{pseudo-rehearsal}). This improves performance because the data from the GAN are more likely to be close to the real distribution of dataset than the sampled real ones. On top of that, the GAN used here resolves the privacy issue by not generating individual-specific data. Although ICwGAN is relatively free from the scalability problem, computation overhead by training and running GAN is inevitable. Gradient Episodic Memory (GEM) \cite{NIPS2017_7225} adopts an external memory like iCaRL, but aims to not only reduce forgetting but also make positive backward transfer possible by using inequality constraints and solving dual problem of a quadratic program with them.

The following works implement \textit{dual memory system} inspired by complementary learning system (CLS) theory \cite{mcclelland1995there,kumaran2016learning} that investigated the interaction between hippocampus and neocortex in mammalian brain. GeppNet \cite{gepperth:hal-01418123} makes use of self-organized map algorithm to obtain topology-preserving representation at the hidden layer; the representation is updated only when current classification with linear regression produces uncertain or wrong result. Thus, the hidden layer can serve as a stable long-term memory. FearNet \cite{kemker2018fearnet} is a recent approach that does not store previous exemplars for retraining. Its long-term memory is modelled by an autoencoder network, which is capable of generating previously learned examples. When new classes are observed, FearNet first stores those observations in its short-term memory, then consolidates those observations into the long-term memory along with generated examples. The short-term memory is erased after consolidation, thus acts as a finite-sized episodic memory.

\section{Methodology}

In this section, we briefly introduce ICL and describe the components of our model, each of which is followed by \textit{background} subsection to explain motivation and details.

\subsection{Incremental Class Learning}

The inputs under ICL setting are given as $X^1,X^2,...,$ where $X^t=\{x^t_1,x^t_2,...,x^t_{N_t}\}$ is a set of data and $Y^t=\{y^t_1,y^t_2,...,y^t_{N^t_c}\}$ is a set of classes present in $X^t$ where $N^t_c$ denotes the number of classes for a task $t$ . $N^{t_1}_c$ is set to a half of total number of classes in the dataset we use, and the other tasks contain the same number of classes of our choice (e.g., with CIFAR100, $N^{t_1}_c$ equals 50, and $N^{t_2}_c,N^{t_3}_c,...,N^{t_{51}}_c$ are all 1). 

An ICL model is required to learn $t_1$, which we call \textit{base training}. Then, it should learn $t_2,t_3,...$ sequentially and we call this as \textit{incremental training}. The classes are disjointedly split among tasks. 

\begin{figure}[h]
\includegraphics[width=0.4\textwidth]{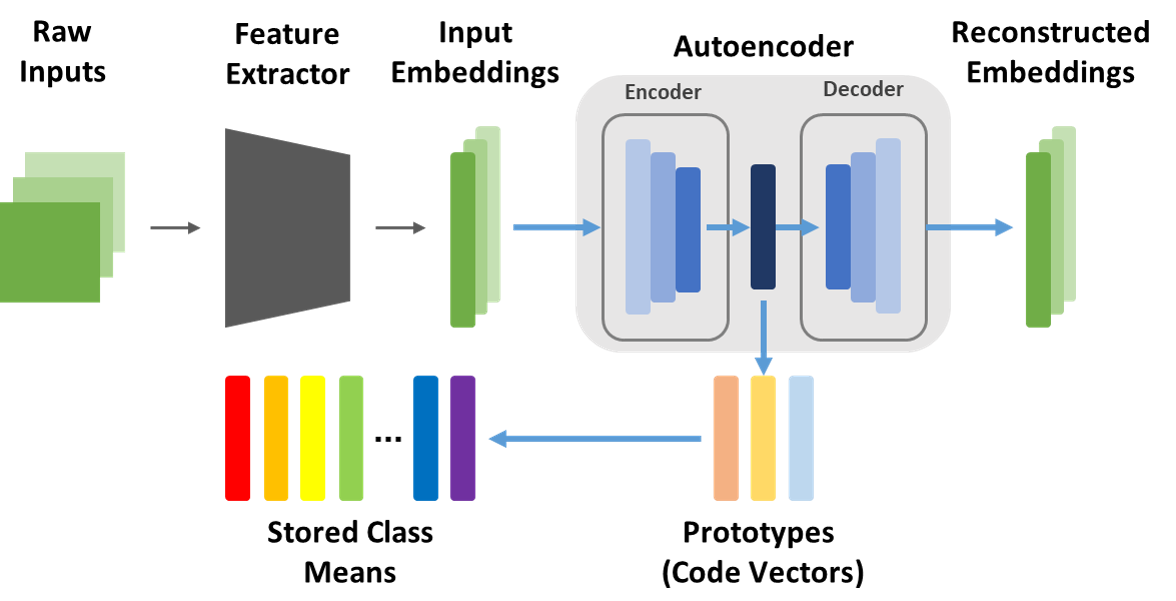}
\centering

\caption{Overall architecture of our model.}
\end{figure}
\subsection{Architecture and Classification}
\vspace{-0.5em}
\subsubsection{Autoencoder-Based Architecture}
We use an autoencoder as a classifier, instead of conventional multi-layer perceptrons (MLP) or convolutional neural network (CNN) architectures trained with the softmax at the final layer. The outputs of the encoder (i.e., the code vectors of the autoencoder) are used as the \textit{prototypes} for given inputs (see Figure 1). Since a fraction of the entire dataset is not sufficient to train the whole network in case of the datasets we experiment on (CIFAR-100 and CUB-200), a VGG-19 \cite{simonyan2014very} pretrained on ImageNet is attached in front of the encoder and works as a fixed feature extractor $\varphi$. This is similar to FearNet, which uses a pretrained ResNet-50 \cite{he2016deep} as its feature extractor.

\subsubsection{Cosine Similarity-Based Classification}

The cosine version of NCM is used for classification. In our model, the predicted class for a given input $x$ is
\vspace{-0.5em}
\begin{equation}
\vspace{-0.5em}
y=\argmin_{i\in\{0,1,...,N_c-1\}} cos(h(x), \mu_i)
\end{equation}
where $0,1,...,N_c-1$ indicate labels for total $N_c$ classes observed so far, $h(\cdot)$ indicates the encoder operation, $\mu_i$ indicates the class mean of prototypes for $i$th class, and $cos(\cdot,\cdot)$ is cosine similarity distance.

Class means are calculated only once after training the autoencoder for each task. That is, whole training data in task $t$ is fed to the autoencoder in mini-batches and the class means are calculated, as proposed by \cite{guerriero2018deepncm:}.

\subsubsection{Background} 
A lot of previous approaches make their prediction by interpreting a network's softmax output as a class probability and this can be problematic for ICL. First, the number of nodes at the output layer always has to grow and the total number of classes may not be known in real-world applications. Second, unlike in multi-task learning, a network has only one growing output layer (i.e., which task a given input belongs to is not known to the network). This makes ICL training harder because the possibility that a network predicts an input from new class as one of old classes is left open \cite{maltoni2018continuous}. In other words, a network has to solve two problems simultaneously for ICL; one for classifying task, the other for predicting correct class.

Autoencoder architecture can readily be used in this context as it does not require softmax output unit, hence free from the aforementioned concerns. The choice of metric-based classification rule is followed naturally because it does not assume a specific number of classes. More importantly, small drifts in the model less affect classification performance, as far as the drifted output remains in the neighborhood of the original output.

\subsection{Loss Function for Base Training}

We propose a loss function to train on base classes which consists of three components. The first term is pixel-by-pixel MSE reconstruction loss commonly used to train an autoencoder, given as:
\vspace{-0.25em}
\begin{equation}
\vspace{-0.25em}
L_{MSE}={\Vert}g(h(x))-x{\Vert}^2_2
\end{equation}
where $h(\cdot)$ and $g(\cdot)$ denote the encoder and the decoder operation, respectively. The second term is cosine embedding loss to make code vectors well separated in cosine metric space according to the class they belong to, which is calculated as:
\vspace{-0.25em}
\begin{equation}
\vspace{-0.25em}
L_{cos}=\begin{cases}
            1-cos(x_1,x_2), & \text{if $y_1=y_2$}\\
            max(0,cos(x_1,x_2)), & \text{if $y_1\neq{y_2}$}
        \end{cases}.
\end{equation}
Though there are $N_b\choose{2}$ many possible pairs for $(x_1,x_2)$, we randomly select $N_b$ many pairs to reduce computation overhead ($N_b$ denotes the mini-batch size).
The last term is L1 penalty loss to make code vectors sparse, thus aiding the cosine embedding loss and is defined as:
\vspace{-0.25em}
\begin{equation}
\vspace{-0.25em}
L_{L1}=\sum_{i} |h_i(x)|
\end{equation}
where $h_i(x)$ means $i$th component of a code vector.

The total loss function to minimize is a weighted linear sum of the three:
\vspace{-0.25em}
\begin{equation}
\vspace{-0.25em}
L_{base}=\lambda_{MSE}L_{MSE}+\lambda_{cos}L_{cos}+\lambda_{L1}L_{L1}
\end{equation}
where the lambdas are hyperparameters to choose.

\subsubsection{Background}
If an autoencoder is trained solely by the reconstruction loss, the prototypes are not guaranteed to be useful for classification tasks since information that is important for imitating every single sample is learned, rather than favorable invariant features to discriminate different classes, as noted by \cite{rasmus2015semi}. This is why we apply the supervised cosine embedding loss term that can bring the prototypes of the same class closer and ones of different classes far away. Plus, adding L1 penalty to the prototypes can further improve classification performance. As sparse vectors lie in narrower regions (that consists of fewer orthants) than dense vectors, the network can concentrate on choosing proper orthants according to the class information, which is simply selecting a few non-zero elements of code vectors. Here, cosine metric gives more advantage due to its magnitude ignorant nature, reducing the degree of freedom of the optimization problem to solve.

\subsection{Outlier Exclusion and Additional Training for Base Classes}
To further increase our model's ability to learn new tasks, we enhance the mean prototypes of base classes via applying Local Outlier Factor (LOF, \cite{breunig2000lof}) and additionally training the encoder to fit to the altered mean prototypes. After base training is complete, the prototypes of all training data are collected class by class, and the outliers in terms of cosine similarity are excluded by LOF to calculate new mean prototypes $\{\mu_{new,i}\}$. Then, the encoder is trained again on the same training data with new loss function consisting of $L_{center}$ and $L_{cos}$, where
\begin{equation}
L_{center}=\sum_{i} {\Vert}h(x)-\mu_{new,i}{\Vert}_2^2
\end{equation}
is a term analogous to the center loss proposed in \cite{wang2018deep} and the total loss function is given as
\begin{equation}
L_{add}=\lambda_{center}L_{center}+\lambda_{cos}L_{cos}
\end{equation}
with the lambdas are hyperparameters to determine each term's strength.

\subsubsection{Background}
We observed that the prototypes after base training were not separated well to show good performance in our cosine similarity based classification rule, thus lowering discriminability of the calculated mean prototypes. A natural solution to this problem is excluding outliers before calculating means, however, applying only the outlier exclusion was not powerful enough to enhance our model's performance. We believe this is because the exclusion technique does not change the autoencoder's representation, so the bad prototypes can still be generated from test data, degrading classification accuracy even though mean prototypes are enhanced. Therefore, an additional training step is required to update the encoder to drag outlier prototypes to the enhanced class means and it appears the center loss is adequate to this task.

\subsection{Loss Function for Incremental Training}

Directly fine-tuning the autoencoder also leads to catastrophic forgetting. Therefore, we adopt regularization techniques (SI and MAS) into our model.

SI computes the contribution (or importance) of a parameter $\theta_k$ to the change in the loss function for a task $t_n$ as follows:
\vspace{-0.25em}
\begin{equation}
\vspace{-0.25em}
\omega^n_k=\int_{t_{n-1}}^{t_n}grad_k(\theta(t))\frac{d\theta_k(t)}{dt}dt
\end{equation}
where $grad_k(\theta(t))$ represents the gradient values of the loss function w.r.t. $\theta_k(t)$, and $\frac{d\theta_k(t)}{dt}$ represents the amount of the parameter update. $\omega^n_k$ is then normalized as:
\vspace{-0.5em}
\begin{equation}
\vspace{-0.5em}
\Omega^n_k=\sum_{n_i<n} \frac{\omega^{n_i}_k}{(\Delta^{n_i}_k)^2+\xi}
\end{equation}
where $\Delta^{n_i}_k$ represents the the difference in parameter values before and after training on a task $t_{n_i}$, that is $\Delta^{n_i}_k=\theta_k(t_{n_i})-\theta_k(t_{n_{i-1}})$, and $\xi$ represents a small nonzero damping parameter to prevent numerical overflow. $\Omega^n_k$ accumulates as more tasks are learned, preserving information from the old tasks. It can be combined with quadratic penalty on the parameter drift to form a loss term:
\vspace{-0.25em}
\begin{equation}
\vspace{-0.25em}
L_{SI}=\sum_{k} \Omega^n_k(\tilde{\theta_k}-\theta_k)^2
\end{equation}
where $\tilde{\theta_k}$ is the reference parameter value from the previous task $t_{n-1}$ and $\theta_k$ is the value immediately after learning a task $t_n$.

MAS focuses on the function $F$ learned by a network, so suggests an alternative importance measure of a parameter $\theta_k$ given as:
\vspace{-0.25em}
\begin{equation}
\vspace{-0.25em}
\Omega^{n}_{k}=\frac{1}{N} \sum_{i=1}^N {\Vert}grad_k(x_i){\Vert}
\end{equation}
where N is the total number of observed data at a task $t_n$ and $grad_k(x_i)=\frac{\partial(F(x_i;\theta))}{\partial\theta_k}$ for an input $x_i$. The loss term of MAS is defined likewise:
\vspace{-0.25em}
\begin{equation}
\vspace{-0.25em}
L_{MAS}=\sum_{k} \Omega^n_{k}(\tilde{\theta_k}-\theta_k)^2
\end{equation}
where $\tilde{\theta_k}$ and $\theta_k$ mean the same as in SI.

The total loss function for learning new tasks is similar to $L_{base}$ except $L_{cos}$ can be excluded according to class split. That is:
\vspace{-0.25em}
\begin{equation}
\vspace{-0.25em}
L_{inc}=\lambda_{MSE}L_{MSE}+\lambda_{reg}L_{reg}+\lambda_{cos}L_{cos}+\lambda_{L1}L_{L1}
\end{equation}
where $L_{reg}$ is one of $L_{SI}$ and $L_{MAS}$.

\subsubsection{Background}
The reason SI and MAS are preferred to EWC is that they are more apt for online updates with small additional computation costs, and more scalable to higher output dimensions (the Fisher information matrix used in EWC is computationally expensive).

Given that ReLU activation is used, we can select $F$ mentioned in MAS to be either the loss function or the activation output of any hidden layer of the network (a \textit{local} version as proposed in \cite{10.1007/978-3-030-01219-9_9}). But we just set $F$ to the total loss function because if a specific hidden layer is selected, it might be biased to a part of tasks the autoencoder has to perform (e.g., the hidden layer of the decoder is likely to be optimized for reconstruction).

\vspace{-0.5em}
\begin{table*}[ht]
\begin{minipage}{0.5\textwidth}
\centering
\small
\begin{tabular}{l|c|c}
\hline
                                & \textbf{CIFAR-100} & \textbf{CUB-200-2011} \\
\hline\hline
Classes                         & 100                & 200              \\ \hline
Train Samples                   & 50,000             & 5,994            \\ \hline
Train Samples / Class           & 500                & 29 to 30         \\ \hline
Test Samples                    & 10,000             & 5,794            \\ \hline
Test Samples / Class            & 100                & 11 to 30         \\
\hline
\end{tabular}
\vspace{-0.5em}
\caption{Dataset statistics}
\vspace{-0.5em}
\end{minipage}
\begin{minipage}{0.5\textwidth}
\small
\begin{tabular}{l|c|c}
\hline
\multicolumn{1}{l|}{}        & \textbf{CIFAR-100}           & \textbf{CUB-200-2011}         \\ \hline\hline
Input Image Dim.             & 3 $\times$ 32 $\times$ 32    & 3 $\times$ 224 $\times$ 224   \\ \hline
Feature Extractor $\varphi$  & To first 9 conv. layers         & To first FC layer                \\ \hline
Input Embedding Dim.         & 8192                         & 4096                          \\ \hline
Prototype Dim.               & 2048                         & 1024                          \\ \hline
Enc./Dec. Architecture & \begin{tabular}[c]{@{}c@{}}{[}8192-2048-2048{]}\\ Activation: ELU\end{tabular} & \begin{tabular}[c]{@{}c@{}}{[}4096-1024-1024{]}\\ Activation: ELU\end{tabular} \\ \hline
\end{tabular}
\vspace{-0.5em}
\caption{Experimental settings for each dataset}
\vspace{-0.5em}
\end{minipage}
\end{table*}

\subsection{Training}
We design procedures for base training and incremental training, respectively. Since the two regularization techniques' requirement differs, we also make this point clear.

\subsubsection{Base Training}
The autoencoder is trained on $t_1$. In an epoch, training dataset $X^1$ is randomly shuffled and divided into mini-batches of size $N_b$. The initial values of autoencoder parameters $\Theta=\{\theta_k\}$ are stored to $\tilde{\Theta}$. For each mini-batch, input images are fed to the feature extractor to obtain embeddings $\varphi(x)$ which are flattened to have shape of $(N_b,-)$. $L_{MSE}(\varphi(x))$ and $L_{L1}(\varphi(x))$ are calculated straightforward. For $L_{cos}$, $N_b$ many pairs of $(\varphi(x_1), \varphi(x_2))$ are randomly sampled ($x_1\neq{x_2}$). If $L_{SI}$ is used for incremental training, both the parameter values before update $\Theta^*$ and gradient values of $L_{base}$ w.r.t. $\Theta^*$ is stored. In case of $L_{MAS}$, only the gradient values are kept. Then we update $\Theta$ via back-propagation and accumulate per-parameter importance values ($\{\omega^1_k\}$ in Eq. 6 for SI, $\{\Omega^1_k\}$ in Eq. 9 for MAS). After the final epoch, the whole training dataset is fed to the autoencoder and we separately collect the output of the encoder $h(\varphi(x))$ class by class to make class mean of prototypes $\{\mu_i\}$. If the outlier exclusion and the additional training is used, LOF is applied to $\{\mu_i\}$ to obtain $\{\mu_{new,i}\}$ and the encoder is trained by minimizing $L_{add}$ in Eq. 7. Here, per-parameter importance values are accumulated in the same way, now w.r.t. $L_{add}$. Then, the final importance value is normalized and then saved. For SI, $\Delta_k$ is computed as $\theta_k-\tilde{\theta_k}$ and $\Omega^1_k$ is computed as in Eq. 9. And for MAS, $\Omega^1_k$ is divided by the total number of inputs in training dataset.

\subsubsection{Incremental Training}
Next, the rest of tasks $t_2,t_3,...$ are learned. Most of the procedures are the same as in base training, however, only the encoder is updated. $L_{base}$ is replaced by $L_{inc}$ and the outlier exclusion and the additional training are not applied because they make the encoder overfit to data from new tasks (only one class included in each task). At training for each task, the importance measure obtained in the last task is used to compute $L_{SI}$ or $L_{MAS}$, and a new importance measure is calculated for the task following right after. When the training is completed, $\{\mu_i\}$ is appended with new class means. It is notable that each $\Omega^{j}_k$ for $j\geq2$ embraces information about parameter importance from all the previous tasks, thus showing implicit effect of Knowledge Distillation without depending on rehearsal or pseudo-rehearsal techniques.

\section{Experiments}

We demonstrate the performance of our model in various ICL settings. In this section, we explain our experimental settings, and then show results along with comparison with notable previous approaches.

\subsection{Experimental Settings}

\subsubsection{Dataset}
We evaluate our model on two datasets in Table 1: \textit{CIFAR-100} \cite{krizhevsky2009learning} and \textit{CUB-200-2011} \cite{WahCUB_200_2011}. \textit{CIFAR-100} is one of the most popular RGB image datasets for classification. The image shape is 3 $\times$ 32 $\times$ 32 pixels and 100 classes of various objects are included. \textit{CUB-200-2011} is a challenging dataset containing 200 species of birds, with much smaller number of examples per class. It also provides additional information for object detection, but we use it solely for classification task. The image shapes are not uniform, so we preprocess the images as follows: 1) resize an image so that the smaller spatial dimension is set to 224, 2) crop at the center to get the final shape as 3 $\times$ 224 $\times$ 224 pixels.

\subsubsection{Implementation Details}
The architectures and settings of our model for each dataset are shown in Table 2. For \textit{CIFAR-100}, the feature map after first 9 convolutional layers of VGG-19 \cite{simonyan2014very} pretrained on ImageNet is used as the fixed feature extractor $\varphi$. For \textit{CUB-200-2011}, the activation output of the first fully-connected layer from the same VGG-19 is used as $\varphi$. For all dataset, our model is trained with AMSGrad \cite{j.2018on} with mini-batches of size 64 and ELU \cite{clevert2015fast} is used as activation function. We implemented our model with PyTorch.

\subsubsection{Evaluation Scheme}
We evaluate our model's performance when each dataset is split into disjoint subsets. Specifically, the first subset contains a half of all classes in a dataset and each other subset contains a single class. For direct comparison with previous works, we use evaluation metrics suggested by \cite{kemker2018measuring}. Three metrics were proposed: $\Psi_{base}$, $\Psi_{new}$, and $\Psi_{all}$ (the original notation $\Omega$ from Kemker et al. is changed to $\Psi$ to eliminate confusion with $\Omega^n_k$ for SI and MAS), where $\Psi_{base}=\frac{1}{T-1}\sum_{i=2}^T \frac{\alpha_{base,i}}{\alpha_{ideal}}$ represents a model's ability to maintain its performance on base classes it first learned (referred to as the "base knowledge"), $\Psi_{new}=\frac{1}{T-1}\sum_{i=2}^T \alpha_{new,i}$ represents a model's ability to perform well on newly learned classes, and $\Psi_{all}=\frac{1}{T-1}\sum_{i=2}^T \frac{\alpha_{all,i}}{\alpha_{ideal}}$ represents a model's overall classification performance on all learned classes. Notations with $\alpha$ stands for accuracy, for example $\alpha_{base,i}$ means the accuracy on the base knowledge after the $i$th training session. $\alpha_{ideal}$ is a regularizing term representing the ideal accuracy when all classes are presented at once (offline setting), thereby meaning the upper bound of model's performance. The value is 0.699 for \textit{CIFAR-100} and 0.598 for \textit{CUB-200-2011}. $T$ is the total number of tasks, which becomes 51 for \textit{CIFAR-100} and 101 for \textit{CUB-200-2011}.

\subsubsection{Model Configurations}
Our model is experimented with 4 different configurations. \textit{SI} and \textit{MAS} mean that $L_{SI}$ and $L_{MAS}$ was used in Eq. 13 for incremental training steps, respectively. In \textit{SI+LOF} and \textit{MAS+LOF}, outlier exclusion and additional training are further applied at the base training, and the rest is the same with \textit{SI} and \textit{MAS}.

\subsection{Results}

\subsubsection{CIFAR-100}
\vspace{-0.5em}
\begin{table}[h]
\centering
\begin{tabular}{l|ccc}
\hline
Model   & $\Psi_{base}$ & $\Psi_{new}$ & $\Psi_{all}$ \\ \hline\hline
1-NN\footnotemark            & 0.878  & 0.648  & 0.879  \\
GeppNet\footnotemark[\value{footnote}]         & 0.833  & 0.529  & 0.754  \\
iCaRL\footnotemark[\value{footnote}]           & 0.746  & 0.807  & 0.749  \\
FearNet\footnotemark[\value{footnote}]         & \textbf{0.927}  & \textbf{0.824}  & \textbf{0.947}  \\ \hline
Ours w/ MAS     & 0.887  & 0.641  & 0.850  \\
Ours w/ SI      & 0.864  & 0.618  & 0.857  \\
Ours w/ MAS+LOF & 0.859  & 0.746  & 0.823  \\
Ours w/ SI+LOF  & 0.833  & 0.697  & 0.814  \\ \hline
\end{tabular}
\vspace{-0.5em}
\caption{Incremental classification metrics on CIFAR-100}
\vspace{-0.5em}
\end{table}

\vspace{-0.5em}
\begin{table}[h]
\centering
\begin{tabular}{l|cc}
\hline
Model   & 100 Classes & 1,000 Classes \\ \hline\hline
1-NN\footnotemark[\value{footnote}]    & 4.1GB       & 40.9GB        \\
GeppNet\footnotemark[\value{footnote}] & 4.1GB       & 41GB          \\
iCaRL   & 24.8MB      & 247.8MB       \\
FearNet & 3.3MB       & 18MB        \\ \hline
Ours    & \textbf{0.8MB}       & \textbf{8.2MB}         \\ \hline
\end{tabular}
\vspace{-0.5em}
\caption{Estimated extra memory requirement values on CIFAR-100. Note that for iCaRL, $K$ (total capacity of stored exemplars) was set to 2,000 for 100 classes and 20,000 for 1,000 classes. And for FearNet, the number of exemplars stored per class $m$ is set to 20 and sleep frequency is 10 epochs.}
\end{table}
\noindent
Table 3 shows the metrics by Kemker et al. measured with \textit{CIFAR-100}. For \textit{Ours w/ MAS} and \textit{Ours w/ SI}, the base knowledge is trained for 100 epochs with fixed learning rate of 1e-4, and the incremental training is done for 50 epochs with fixed learning rate of 2e-4 and 10 epochs with fixed learning rate of 1e-4, respectively. For \textit{Ours w/ MAS+LOF} and \textit{Ours w/ SI+LOF}, the base knowledge is trained for 100 epochs with learning rate decay of 0.2 for every 25 epochs, starting with 2e-4, and the incremental training is done for 50 epochs with fixed learning rate of 1e-4 and 25 epochs with fixed learning rate of 2e-4, respectively. Hyperparameter settings are consistent among all configurations, with $\lambda_{MSE}$=1, $\lambda_{cos}$=10, $\lambda_{L1}$=1e-3, $\lambda_{reg}$=10, and $\lambda_{center}$=1.

Our model, especially \textit{Ours w/ MAS}, shows competitive incremental classification results, with $\Psi_{all}$ slightly worse than 1-NN, but better than GeppNet and iCaRL. While they must explicitly store previous examples, we achieve better or comparable incremental classification ability with storing only one mean prototype per class. Also, our model is better at retaining knowledge for base classes compared to all three of them as shown in $\Psi_{base}$ values, and this explains the good $\Psi_{all}$ values of our model despite relatively low $\Psi_{new}$ (to iCaRL); although our model might not immediately show good performance on the most recent class, it forgets less as the learning progresses. Applying LOF increases $\Psi_{new}$ with slightly decreased $\Psi_{base}$ and $\Psi_{all}$, making our model's ability to learn new class closer to iCaRL. 

However, our model does not perform as well as FearNet. FearNet essentially emulates joint training with generated training examples from previous classes when its generator (decoder) is perfect, which explains its higher performance compared to models that only rely on new examples like ours, or that use a fraction of previous examples instead of full joint training like iCaRL. Thus, although FearNet has higher performance, its training cost and time scale with the number of learned classes. Furthermore, FearNet's performance depends on the frequency of "sleep" (consolidation) phase, and for better $\Psi_{all}$, data from more classes should be stored in its short-term memory before consolidation happens, which further increases memory cost. On the other hand, our model's training cost and time is constant regardless of the number of classes observed so far because only the data from a new class is used for every task. Moreover, our model's memory requirement grows much slower than FearNet's since whatever configuration we choose, the number of stored mean prototypes is the same as the number of classes. This point is made clear in Table 4 where the estimated extra memory requirement after learning 100 and 1,000 classes are shown. To calculate the values, we included all additional data or statistics required to perform the final incremental training step given the number of classes, such as stored exemplars, class means, or variances.

\subsubsection{CUB-200-2011}
\begin{table}[h]
\centering
\begin{tabular}{l|ccc}
\hline
Model   & $\Psi_{base}$ & $\Psi_{new}$ & $\Psi_{all}$ \\ \hline\hline
1-NN\footnotemark[\value{footnote}]            & 0.746  & 0.434  & 0.694  \\
GeppNet\footnotemark[\value{footnote}]         & 0.727  & 0.558  & 0.645  \\
iCaRL\footnotemark[\value{footnote}]           & 0.942  & 0.547  & 0.864  \\
FearNet\footnotemark[\value{footnote}]         & \textbf{0.924}  & 0.598  & \textbf{0.891}  \\ \hline
Ours w/ MAS+LOF & 0.836  & \textbf{0.780}  & 0.727  \\
Ours w/ MAS     & 0.837  & 0.672  & 0.769  \\
Ours w/ SI      & 0.824  & 0.726  & 0.762  \\
Ours w/ SI+LOF  & 0.813  & 0.639  & 0.737  \\ \hline
\end{tabular}
\vspace{-0.5em}
\caption{Incremental classification metrics on CUB-200-2011}
\vspace{-0.5em}
\end{table}
\footnotetext{The numbers are from \cite{kemker2018fearnet}}
Table 5 shows the same metric in Table 3 measured with \textit{CUB-200-2011}. For \textit{Ours w/ MAS} and \textit{Ours w/ SI}, the base knowledge is trained for 100 epochs with learning rate decay of 0.5 every 25 epochs, starting with 2e-4, and the incremental training is done for 50 epochs with fixed learning rate of 2e-4. For \textit{Ours w/ MAS+LOF} and \textit{Ours w/ SI+LOF}, the base knowledge is trained for 100 epochs with learning rate decay of 0.2 for every 25 epochs, starting with 2e-4, and the incremental training is done for 50 epochs with fixed learning rate of 2e-4 and 25 epochs with fixed learning rate of 2e-4, respectively. Hyperparameter settings are the same as in \textit{CIFAR-100}.

$\Psi_{base}$ and $\Psi_{all}$ of our model are lower than iCaRL and FearNet possibly due to the same reason argued in \textbf{CIFAR-100} subsection. However, in general, our model (especially \textit{Ours w/ MAS+LOF}) shows significantly higher $\Psi_{new}$ than any other model with much smaller and scalable extra memory requirement as can be inferred from Table 4 (i.e., the dimension of class mean prototype is decreased to 1,024 and the number of them is doubled to 200 classes, thus requiring the same amount of extra memory as in \textit{CIFAR-100} whereas the others' memory has to increase due to higher input dimension).

\section{Conclusion}
We propose a novel ICL algorithm that leverages 1) the autoencoder architecture to extract prototypes of inputs, 2) metric-based classification rule (the cosine version of NCM), 3) loss functions for base and incremental training, 4) outlier exclusion and additional training to enhance our model's ability to learn new class, and 5) regularization methods to mitigate catastrophic forgetting (SI and MAS). We demonstrate the performance of our model in various experimental settings, involving different datasets and model configurations. Furthermore, our model does not rely on stored or generated examples, providing memory-efficient approach to ICL. This is possible as regularization method prevent the model from drastic semantic drift, then small changes in the output of the encoder after incremental training steps do not hurt classification accuracy as far as the output has same near neighbors as before in terms of cosine similarity, while not relying on rehearsal or pseudo-rehearsal technique that causes more memory overhead. We believe future work on extending representation learning to complex datasets instead of transferring a pretrained model can improve the performance, because features pretrained on a different dataset might not be optimal to the given dataset at hand.

\bibliographystyle{named}
{\footnotesize\bibliography{ijcai19}}

\begin{thebibliography}{}

\bibitem[\protect\citeauthoryear{Aljundi \bgroup \em et al.\egroup
  }{2018}]{10.1007/978-3-030-01219-9_9}
Rahaf Aljundi, Francesca Babiloni, Mohamed Elhoseiny, Marcus Rohrbach, and
  Tinne Tuytelaars.
\newblock Memory aware synapses: Learning what (not) to forget.
\newblock In {\em Computer Vision -- ECCV 2018}, pages 144--161, 2018.

\bibitem[\protect\citeauthoryear{Breunig \bgroup \em et al.\egroup
  }{2000}]{breunig2000lof}
Markus~M Breunig, Hans-Peter Kriegel, Raymond~T Ng, and J{\"o}rg Sander.
\newblock Lof: identifying density-based local outliers.
\newblock In {\em ACM sigmod record}, volume~29, pages 93--104. ACM, 2000.

\bibitem[\protect\citeauthoryear{Clevert \bgroup \em et al.\egroup
  }{2015}]{clevert2015fast}
Djork-Arn{\'e} Clevert, Thomas Unterthiner, and Sepp Hochreiter.
\newblock Fast and accurate deep network learning by exponential linear units
  (elus).
\newblock {\em arXiv preprint arXiv:1511.07289}, 2015.

\bibitem[\protect\citeauthoryear{French}{1999}]{french1999catastrophic}
Robert~M French.
\newblock Catastrophic forgetting in connectionist networks.
\newblock {\em Trends in cognitive sciences}, 3(4):128--135, 1999.

\bibitem[\protect\citeauthoryear{Gepperth and
  Karaoguz}{2016}]{gepperth:hal-01418123}
Alexander Gepperth and Cem Karaoguz.
\newblock {A Bio-Inspired Incremental Learning Architecture for Applied
  Perceptual Problems}.
\newblock {\em {Cognitive Computation}}, 8:924 -- 934, 2016.

\bibitem[\protect\citeauthoryear{Guerriero \bgroup \em et al.\egroup
  }{2018}]{guerriero2018deepncm:}
Samantha Guerriero, Barbara Caputo, and Thomas Mensink.
\newblock Deepncm: Deep nearest class mean classifiers.
\newblock In {\em International Conference on Learning Representations,
  Workshop Track}, 2018.

\bibitem[\protect\citeauthoryear{He \bgroup \em et al.\egroup
  }{2016}]{he2016deep}
Kaiming He, Xiangyu Zhang, Shaoqing Ren, and Jian Sun.
\newblock Deep residual learning for image recognition.
\newblock In {\em Computer Vision and Pattern Recognition (CVPR)}, pages
  770--778. IEEE, 2016.

\bibitem[\protect\citeauthoryear{Hinton \bgroup \em et al.\egroup
  }{2015}]{44873}
Geoffrey Hinton, Oriol Vinyals, and Jeffrey Dean.
\newblock Distilling the knowledge in a neural network.
\newblock In {\em NIPS Deep Learning and Representation Learning Workshop},
  2015.

\bibitem[\protect\citeauthoryear{Hsu \bgroup \em et al.\egroup
  }{2018}]{hsu2018re}
Yen-Chang Hsu, Yen-Cheng Liu, and Zsolt Kira.
\newblock Re-evaluating continual learning scenarios: A categorization and case
  for strong baselines.
\newblock {\em arXiv preprint arXiv:1810.12488}, 2018.

\bibitem[\protect\citeauthoryear{Kemker and Kanan}{2018}]{kemker2018fearnet}
Ronald Kemker and Christopher Kanan.
\newblock Fearnet: Brain-inspired model for incremental learning.
\newblock In {\em International Conference on Learning Representations}, 2018.

\bibitem[\protect\citeauthoryear{Kemker \bgroup \em et al.\egroup
  }{2018}]{kemker2018measuring}
Ronald Kemker, Marc McClure, Angelina Abitino, Tyler~L Hayes, and Christopher
  Kanan.
\newblock Measuring catastrophic forgetting in neural networks.
\newblock In {\em Thirty-Second AAAI Conference on Artificial Intelligence},
  2018.

\bibitem[\protect\citeauthoryear{Kirkpatrick \bgroup \em et al.\egroup
  }{2017}]{kirkpatrick2017overcoming}
James Kirkpatrick, Razvan Pascanu, Neil Rabinowitz, Joel Veness, Guillaume
  Desjardins, Andrei~A Rusu, Kieran Milan, John Quan, Tiago Ramalho, Agnieszka
  Grabska-Barwinska, et~al.
\newblock Overcoming catastrophic forgetting in neural networks.
\newblock {\em Proceedings of the national academy of sciences}, page
  201611835, 2017.

\bibitem[\protect\citeauthoryear{Krizhevsky and
  Hinton}{2009}]{krizhevsky2009learning}
Alex Krizhevsky and Geoffrey Hinton.
\newblock Learning multiple layers of features from tiny images.
\newblock Technical report, Citeseer, 2009.

\bibitem[\protect\citeauthoryear{Kumaran \bgroup \em et al.\egroup
  }{2016}]{kumaran2016learning}
Dharshan Kumaran, Demis Hassabis, and James~L McClelland.
\newblock What learning systems do intelligent agents need? complementary
  learning systems theory updated.
\newblock {\em Trends in cognitive sciences}, 20(7):512--534, 2016.

\bibitem[\protect\citeauthoryear{Li and Hoiem}{2018}]{li2018learning}
Zhizhong Li and Derek Hoiem.
\newblock Learning without forgetting.
\newblock {\em IEEE Transactions on Pattern Analysis and Machine Intelligence},
  40(12):2935--2947, 2018.

\bibitem[\protect\citeauthoryear{Lopez-Paz and Ranzato}{2017}]{NIPS2017_7225}
David Lopez-Paz and Marc~Aurelio Ranzato.
\newblock Gradient episodic memory for continual learning.
\newblock In {\em Advances in Neural Information Processing Systems 30}, pages
  6467--6476. Curran Associates, Inc., 2017.

\bibitem[\protect\citeauthoryear{Maltoni and
  Lomonaco}{2018}]{maltoni2018continuous}
Davide Maltoni and Vincenzo Lomonaco.
\newblock Continuous learning in single-incremental-task scenarios.
\newblock {\em arXiv preprint arXiv:1806.08568}, 2018.

\bibitem[\protect\citeauthoryear{McClelland \bgroup \em et al.\egroup
  }{1995}]{mcclelland1995there}
James~L McClelland, Bruce~L McNaughton, and Randall~C O'reilly.
\newblock Why there are complementary learning systems in the hippocampus and
  neocortex: insights from the successes and failures of connectionist models
  of learning and memory.
\newblock {\em Psychological review}, 102(3):419, 1995.

\bibitem[\protect\citeauthoryear{McCloskey and
  Cohen}{1989}]{mccloskey1989catastrophic}
Michael McCloskey and Neal~J Cohen.
\newblock Catastrophic interference in connectionist networks: The sequential
  learning problem.
\newblock In {\em Psychology of learning and motivation}, volume~24, pages
  109--165. 1989.

\bibitem[\protect\citeauthoryear{Mensink \bgroup \em et al.\egroup
  }{2013}]{6517188}
T.~Mensink, J.~Verbeek, F.~Perronnin, and G.~Csurka.
\newblock Distance-based image classification: Generalizing to new classes at
  near-zero cost.
\newblock {\em IEEE Transactions on Pattern Analysis and Machine Intelligence},
  35(11):2624--2637, Nov 2013.

\bibitem[\protect\citeauthoryear{Parisi \bgroup \em et al.\egroup
  }{2018}]{parisi2018continual}
German~I Parisi, Ronald Kemker, Jose~L Part, Christopher Kanan, and Stefan
  Wermter.
\newblock Continual lifelong learning with neural networks: A review.
\newblock {\em arXiv preprint arXiv:1802.07569}, 2018.

\bibitem[\protect\citeauthoryear{Rasmus \bgroup \em et al.\egroup
  }{2015}]{rasmus2015semi}
Antti Rasmus, Mathias Berglund, Mikko Honkala, Harri Valpola, and Tapani Raiko.
\newblock Semi-supervised learning with ladder networks.
\newblock In {\em Advances in Neural Information Processing Systems}, pages
  3546--3554, 2015.

\bibitem[\protect\citeauthoryear{Rebuffi \bgroup \em et al.\egroup
  }{2017}]{Rebuffi_2017_CVPR}
Sylvestre-Alvise Rebuffi, Alexander Kolesnikov, Georg Sperl, and Christoph~H.
  Lampert.
\newblock icarl: Incremental classifier and representation learning.
\newblock In {\em The IEEE Conference on Computer Vision and Pattern
  Recognition (CVPR)}, July 2017.

\bibitem[\protect\citeauthoryear{Reddi \bgroup \em et al.\egroup
  }{2018}]{j.2018on}
Sashank~J. Reddi, Satyen Kale, and Sanjiv Kumar.
\newblock On the convergence of adam and beyond.
\newblock In {\em International Conference on Learning Representations}, 2018.

\bibitem[\protect\citeauthoryear{Serra \bgroup \em et al.\egroup
  }{2018}]{serra2018overcoming}
Joan Serra, D{\'\i}dac Sur{\'\i}s, Marius Miron, and Alexandros Karatzoglou.
\newblock Overcoming catastrophic forgetting with hard attention to the task.
\newblock {\em arXiv preprint arXiv:1801.01423}, 2018.

\bibitem[\protect\citeauthoryear{Simonyan and
  Zisserman}{2014}]{simonyan2014very}
Karen Simonyan and Andrew Zisserman.
\newblock Very deep convolutional networks for large-scale image recognition.
\newblock {\em arXiv preprint arXiv:1409.1556}, 2014.

\bibitem[\protect\citeauthoryear{Wah \bgroup \em et al.\egroup
  }{2011}]{WahCUB_200_2011}
C.~Wah, S.~Branson, P.~Welinder, P.~Perona, and S.~Belongie.
\newblock {The Caltech-UCSD Birds-200-2011 Dataset}.
\newblock Technical Report CNS-TR-2011-001, California Institute of Technology,
  2011.

\bibitem[\protect\citeauthoryear{Wang \bgroup \em et al.\egroup
  }{2018}]{wang2018deep}
Shuai Wang, Zili Huang, Yanmin Qian, and Kai Yu.
\newblock Deep discriminant analysis for i-vector based robust speaker
  recognition.
\newblock {\em arXiv preprint arXiv:1805.01344}, 2018.

\bibitem[\protect\citeauthoryear{Wu \bgroup \em et al.\egroup
  }{2018}]{wu2018incremental}
Yue Wu, Yinpeng Chen, Lijuan Wang, Yuancheng Ye, Zicheng Liu, Yandong Guo,
  Zhengyou Zhang, and Yun Fu.
\newblock Incremental classifier learning with generative adversarial networks.
\newblock {\em arXiv preprint arXiv:1802.00853}, 2018.

\bibitem[\protect\citeauthoryear{Zenke \bgroup \em et al.\egroup
  }{2017}]{pmlr-v70-zenke17a}
Friedemann Zenke, Ben Poole, and Surya Ganguli.
\newblock Continual learning through synaptic intelligence.
\newblock In {\em Proceedings of the 34th International Conference on Machine
  Learning}, volume~70 of {\em Proceedings of Machine Learning Research}, pages
  3987--3995. PMLR, 06--11 Aug 2017.

\end{thebibliography}

\end{document}